\documentclass{article}

\usepackage{PRIMEarxiv}

\usepackage[utf8]{inputenc} 
\usepackage[T1]{fontenc}    
\usepackage{hyperref}       
\usepackage{url}            
\usepackage{booktabs}       
\usepackage{amsfonts}       
\usepackage{nicefrac}       
\usepackage{microtype}      
\usepackage{lipsum}
\usepackage{fancyhdr}       
\usepackage{graphicx}       
\graphicspath{{media/}}     

\usepackage[backend=biber, maxcitenames=1,maxbibnames=9, minbibnames=5, sorting=none]{biblatex}

\usepackage[symbol]{footmisc}

\usepackage{caption}
\title{a real-time spatiotemporal ai model analyzes skill in open surgical videos}

\newcommand*\samethanks[1][\value{footnote}]{\footnotemark[#1]}

\author{
  Emmett D. Goodman$^{1,2}$\thanks{These authors contributed equally.}, Krishna K. Patel$^{1,2}\samethanks$, Yilun Zhang$^{5}$, William Locke$^{1,2}$, Chris J. Kennedy$^{5,6}$, \\
  \textbf{Rohan Mehrotra$^{1,2}$, Stephen Ren$^{1,2}$, Melody Guan$^{1,2}$, Maren Downing$^{5}$, Hao Wei Chen$^{5}$, Jevin Z. Clark,}$^{5}$  \\
  \textbf{Gabriel A. Brat$^{5,6}$\thanks{Corresponding authors. Emails: gbrat@bidmc.harvard.edu, syyeung@stanford.edu.}, Serena Yeung$^{1,2,3,4}$\samethanks} \\
    $^1$Department of Computer Science, Stanford University, Stanford, CA, USA. \\
    $^2$Department of Biomedical Data Science, Stanford University, Stanford, CA, USA \\
    $^3$Department of Electrical Engineering, Stanford University, Stanford, CA, USA \\
    $^4$Clinical Excellence Research Center, Stanford University School of Medicine, Stanford, CA, USA. \\
    $^5$Department of Surgery, Beth Israel Deaconess Medical Center, Boston, MA, USA. \\
    $^6$Department of Biomedical Informatics, Harvard Medical School, Boston, MA, USA. \\
}

\begin{document}
\maketitle

\begin{abstract}
Open procedures represent the dominant form of surgery worldwide. Artificial intelligence (AI) has the potential to optimize surgical practice and improve patient outcomes, but efforts have focused primarily on minimally invasive techniques. Our work overcomes existing data limitations for training AI models by curating, from YouTube, the largest dataset of open surgical videos to date: 1997 videos from 23 surgical procedures uploaded from 50 countries. Using this dataset, we developed a multi-task AI model capable of real-time understanding of surgical behaviors, hands, and tools—the building blocks of procedural flow and surgeon skill. We show that our model generalizes across diverse surgery types and environments. Illustrating this generalizability, we directly applied our YouTube-trained model to analyze open surgeries prospectively collected at an academic medical center and identified kinematic descriptors of surgical skill related to efficiency of hand motion. Our Annotated Videos of Open Surgery (AVOS) dataset and trained model will be made available for further development of surgical AI.
\end{abstract}


\section{Introduction}
Surgery offers the potential to treat and cure many diseases, but complications from surgical procedures remain the third highest cause of death globally \cite{nepogodiev}. Recent studies have shown that surgeons rated as higher-skilled via peer grading have lower rates of complications and death \cite{sacks, birkmeyer}. Systems to evaluate surgical skill and provide feedback to improve technique could have a dramatic effect on the variation that exists in the field. Unfortunately, current approaches for evaluating surgical procedures and technique are primarily qualitative and do not have the ability to scale or even identify the elements of surgeon judgment that drive patient outcomes \cite{martin}. 

Artificial intelligence (AI) in the form of computer vision algorithms could provide scalable, automated analysis of surgical behaviors from video streams. AI could serve as an additional coach for surgical trainees and as an expert colleague for experienced surgeons \cite{hashimoto}. However, the development of computer vision for open surgery—the dominant form of surgery defined as traditional, non-camera-based surgical techniques \cite{richards}—has been limited by two factors: the complexity of the AI task, and a lack of diverse and sizable training datasets \cite{esteva1}. Our work shows that a multi-task, spatiotemporal AI model, trained on multi-institutional data from numerous surgeons, has the potential to provide consistent analysis and feedback without the bias of any particular surgeon’s experience. 

AI capabilities in the medical domain have seen major advances in computer vision tasks ranging from 2D image classification \cite{esteva2, defauw} to 3D image segmentation \cite{lundervold}. Recent works have also demonstrated success in analyzing echocardiogram videos to predict cardiac ejection fraction \cite{ouyang}. Compared to these data domains, open surgical videos present a greater challenge for computer vision due to the varied number of surgeons, intricate movements of hands and tools, and different operative environments and lighting conditions. Clinically, open surgery is often required for difficult or dangerous surgical procedures, and, because they are not constrained to conditions supporting minimally-invasive technology, often contain countless distinct visual scenarios. For effective analysis of open surgery, new AI methods must be developed which can describe both what is occurring and when, in varied visual scenes. Ideally, such AI models should operate in real time to have greatest utility in the operating room.

To train such a robust model, a diverse AI-ready dataset is key. Dataset curation is simpler for minimally-invasive—or camera-based—surgical procedures; the use of an in-body fiber-optic camera facilitates rapid data collection. In contrast, it is not yet standard practice to record open surgery procedures. Datasets annotated for training AI models are small and are often filmed in bench-top simulations with specialized gloves which limits translation to clinical environments \cite{zia}. In comparison, a widely-used non-surgical video dataset for AI model development contains up to 1000 examples per category of everyday human behaviors \cite{carreira}. A sizable and diverse training dataset is needed to create AI tools which can adapt to the diverse surgical scenes present in open procedures.

\section{The Annotated Videos of Open Surgery (AVOS) Dataset}
\label{sec:headings}

\begin{figure}[h]
    \centering
    \includegraphics[width=0.9\textwidth]{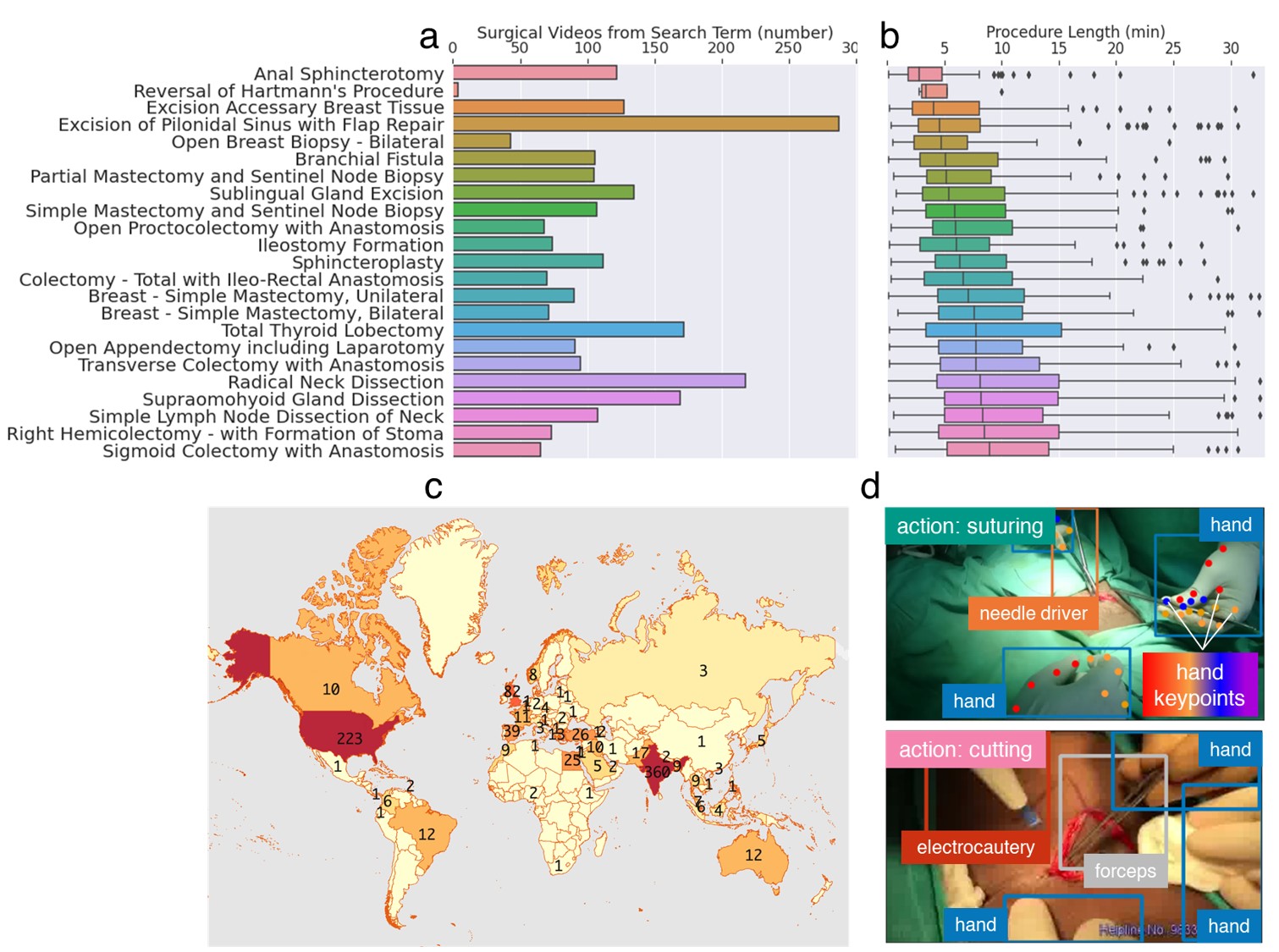}
    \caption{\textbf{Figure 1. Diversity of the AVOS dataset.} Distributions of (a) video counts (b) video duration (only showing data for videos less than 35 min) and (c) number of videos per country (when available). (d) Representative dataset annotations; colored boxes represent hands and tools, points represent hand keypoints, and actions are annotated at the top left of each frame.}
\end{figure}

In recent years, YouTube has emerged as an accessible platform for surgeons and trainees to study surgical videos \cite{chen, derakhshan, rapp}. Our work harnesses YouTube as a massive and diverse video source to train robust AI models that understand surgical scenes. We programmatically queried 23 common open surgery procedures on YouTube, and from 9197 returned results, 1997 open surgical videos were manually identified to form the Annotated Videos of Open Surgery (AVOS) dataset. Importantly, our dataset includes examples from common and rare procedures (Fig. 1a). Videos describing excision or biopsy were on average 4 minutes long, while videos describing more complex procedures, such as those involving colectomy or anastomoses, were on average 8 minutes (Fig. 1b) \cite{deSantibañes}. These videos were uploaded from over 50 countries over the last 15 years, and therefore represent a diverse and international video database (Fig. 1c).

343 videos were annotated with detailed spatial and temporal labels for training computer vision AI models. For each video, ten frames were annotated with rectangular bounding boxes for surgical tools including electrocautery, needle drivers, and forceps. Hands were localized with bounding boxes and 21 joint keypoints (Extended Data Fig. 1). Each video was labeled at a per-second resolution with one of three actions—cutting, tying, or suturing—or a background class including non-surgical content. Representative frame annotations are shown in Fig. 1d and Extended Data Fig. 2. Detailed annotation statistics are in Extended Data Table 1. AVOS represents the largest organized repository of open surgical videos to date, with ~25 times more videos than previous datasets1 \cite{zia}, 47 hours of temporally-segmented surgical content, over 12,000 annotated tools and hands, and over 11,000 identified hand keypoints. 

\section{Detailed Scene Understanding}
\label{sec:headings}

\begin{figure}[h]
    \centering
    \includegraphics[width=0.9\textwidth]{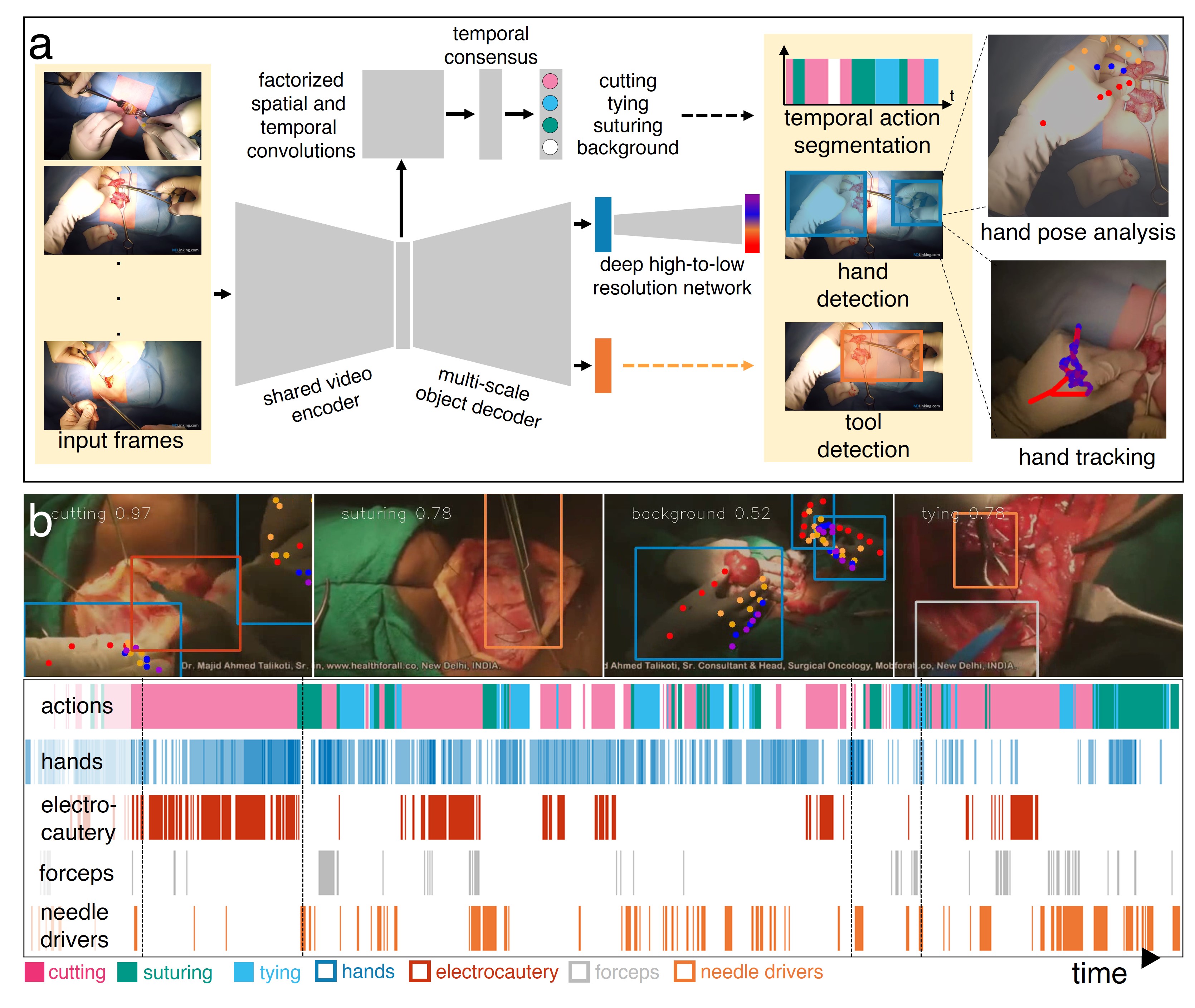}
    \caption{\textbf{Figure 2. Detailed scene understanding using a multi-task neural network.} (a) Sequential video frames pass through a three-headed multi-task neural network for detection of ongoing actions, tools, hands, and hand pose keypoints. Hand detections are fed to the SORT tracking algorithm for movement analysis. (b) Example analysis on an unseen thyroidectomy. Rectangular boxes in each frame represent hand and tool detections, points represent hand pose keypoints, and the long bars below display the progression of surgical actions, hands, and tool detections throughout the procedure.}
\end{figure}

To develop AI capable of comprehensive surgical scene analysis, detailed recognition of both spatial and temporal components is key. We present a multi-task computer vision neural network capable of rapid spatiotemporal inference (Fig. 2a). Through a shared encoding backbone (ResNet-50), we generate a deep scene representation useful across diverse vision tasks. The model then splits into an action recognition branch based on the R(2+1)D architecture with factorized spatial and temporal convolutions \cite{tran}, and two task-specific hand and tool detection branches based on the RetinaNet architecture using a shared Feature Pyramid Network (FPN) for multi-scale object detection \cite{lin}. Action recognition was trained with a cross entropy loss, while focal losses focusing on hard misclassified examples were used for both object detection heads. Hand detections flow into a Deep High-to-Low Resolution Network which utilizes high-resolution image representations to predict 21 hand pose keypoints \cite{sun}. To perform simultaneous optimization of spatial and temporal model branches, we developed a strategy of alternating task training. Alternating batches of short 5s video clips, and randomly sampled images, were used to train the action recognition branch, and hand and tool recognition branches, respectively. Gradients only flowed through the targeted heads and shared encoder at each iteration of training. Through this alternating training scheme, our model converged to a set of parameters which allows for simultaneous processing of video streams for both spatial and temporal tasks. Model detections can be tracked across time using Simple, Online, Real-time Tracking (SORT) \cite{bewley}. Through three parallel analytical pathways in a unified, multi-task neural network architecture, this model understands actions, hands, and tools in diverse surgical videos.

Across the entire AVOS test set of diverse videos with varied qualities and filming angles, the model achieves 0.71 mean precision and 0.73 mean recall for action recognition at one-second temporal resolution. For hand and tool bounding box detection, the model achieves mean average precisions (mAPs) of 0.89 and 0.46, respectively, at an IoU threshold of 0.5. On good quality videos, model performance improves, achieving 0.74 mean precision and 0.79 mean recall for action recognition, and mAPs for hand and tool detection of 0.92 and 0.49. Keypoints achieve an average probability of correct keypoint (PCK) value of 0.38 across all diverse test videos and 0.41 on good quality videos. PCK values are higher for keypoints on the thumb and index fingers, which are key to instrument handling: 0.39 and 0.61 across all test videos, and 0.45 and 0.63 on good quality videos. Additional measures of model performance can be found in Extended Data Tables 2-4. Overall, we observe that the model performs best on high-quality videos at hand-level zoom, suggesting that the most effective clinical application of this tool would occur in such controlled filming environments. Further interpretability of the multi-task model is presented in Extended Data Fig. 3, which shows that pixels close to the objects of interest were key for the detection of hands and instruments. 

As a demonstration of what is possible when a multi-task model performs simultaneous identification of surgical behaviors, hands, and tools, we present an analysis of an open thyroidectomy—a surgery to remove part of an enlarged thyroid gland (Fig. 2b). Detections superimposed on representative frames are shown above, and the identification of actions, hands and tools across time are represented as bar plots below. The early parts of the procedure are characterized by longer periods of cutting, while the middle and end involve alternating episodes of cutting, suturing and tying. Electrocautery device detections (dark red) are strongly correlated with the cutting action, suggesting cutting occurs with an electrocautery rather than another tool—thyroid surgeons often use a cautery to prevent bleeding from tiny blood vessels in the region. The parameterization of a surgery into action and tool sequences serves as a general language that can be used to describe almost all surgical procedures, allowing for nuanced comparison between surgeries and surgeons using the same descriptive building blocks.

\section{Surgical Signatures}
\label{sec:headings}

\begin{figure}[h]
    \centering
    \includegraphics[width=0.9\textwidth]{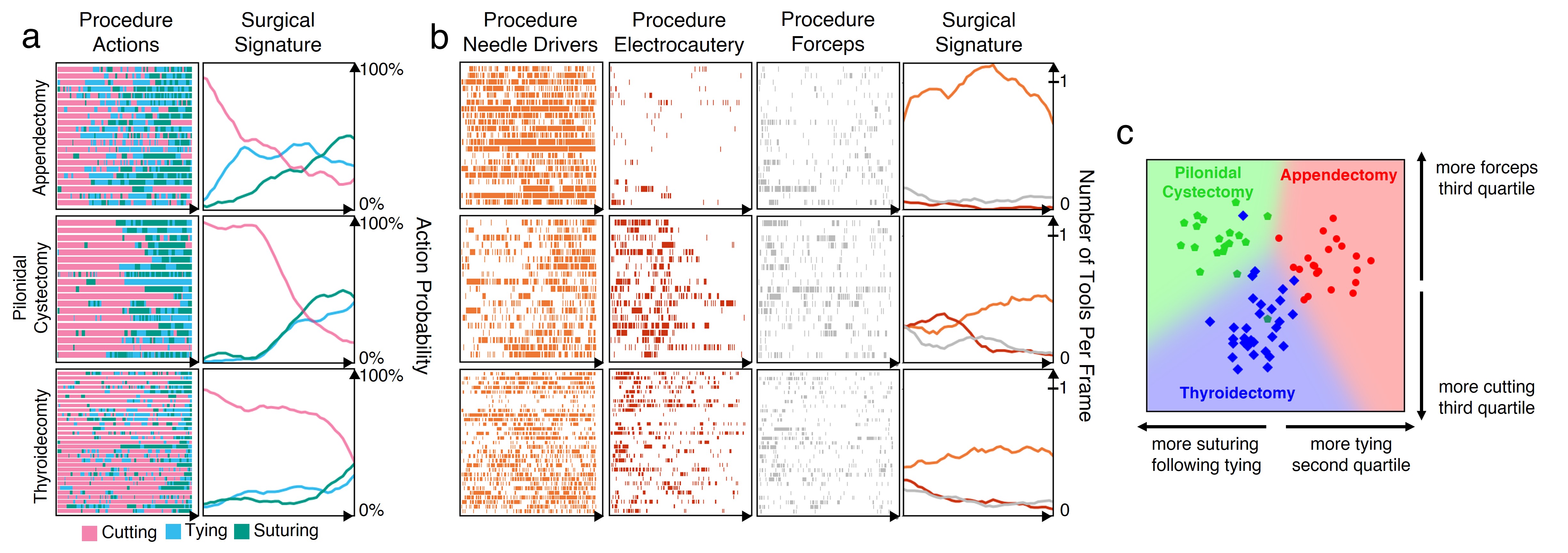}
    \caption{\textbf{Figure 3. Characterizing the structure of diverse open surgical procedures.} (a) Action sequences output from the multi-task neural network, describing individual procedures (left column, each row is a distinct YouTube video), and computed aggregate surgical signature (right column). Background segments were removed for analyses. (b) Tool sequences describing individual procedures (left three columns) and aggregate surgical signature (right column). (c) Identifying key features of each class of surgical procedure through parameterization and dimensionality reduction. The most important features for projection are denoted on each axis.}
\end{figure}

Our multi-task neural network can automatically identify surgical building blocks to build a record of maneuvers and decisions throughout a surgery. Important past work in the development of surgical AI has successfully identified surgical instruments or high-level phases of specific minimally-invasive surgeries, when trained on many examples of the procedure \cite{hashimoto2, garcia, twinanda, ahmidi}. The combination of elemental actions produced by our multi-task network can be used to describe surgical phases in a general language. Our model was trained on diverse surgical videos and techniques from around the world, which allows for analysis across multiple procedure types and surgeon idiosyncrasy. Through interpretation of numerous example cases for a given operation, the model can create a “surgical signature” for each procedure that is robust to the differences between individual surgeons. Clinically, this dense, quantitative description can be used to evaluate surgical skill and technique against a standard reference, and to identify surgeon and procedural anomalies associated with patient outcomes. This objective timeline does not depend on—and is only enriched by—the retrospective dictated note provided by the surgeon. When combined with rapid analysis pipelines (vide infra), this form of AI can ultimately operate as part of a real-time surgical feedback system.

To illustrate the generality of our model, we sorted AVOS by procedure type, and identified open appendectomy, pilonidal cystectomy, and thyroidectomy cases. These videos were extracted by filtering via Unified Medical Language System (UMLS) language tags curated for each video \cite{bodenreider}. Each video was previously unseen by the model, demonstrating its ability to generalize to unseen open surgeries at scale. Using our spatiotemporal multi-task model, we characterized the structure of each surgery, parameterized into a sequence of action and tool appearances. 

Despite a variety of recording angles and video lightings (Extended Data Fig. 4), we can identify latent similarities for each procedure. Fig. 3a, left, shows analysis of actions sequences for each individual video and Fig. 3a, right, shows the corresponding aggregated “surgical signature” created by averaging across procedures. Each action component of the surgical signature represents the probability of that action throughout the procedure. Even though the multi-task model was given no global supervision regarding action sequences, our model finds all surgical signatures begin with an opening incision, evidenced by the near 100\% cutting probability at the beginning of all surgeries. Fig. 3b shows analogous analysis of tool sequences and aggregate signatures. Deviation from a surgical signature, or prototype, may represent impediments to “flow of operation”, which is a known qualitative descriptor of individual surgical skill \cite{martin}, or may indicate an anomalous case. For example, extensive cutting or use of electrocautery during the latter portion of a pilonidal cystectomy may indicate a particularly difficult case with the potential for postoperative complications. 

We parameterized each procedure into 30 features extracted from the multi-task model, including the distribution of actions and tools throughout the procedure (Extended Data Fig. 5). Using those features, we projected each surgery onto two dimensions using linear discriminant analysis. We observe clear separation between the classes of surgeries, indicating that the neural network can create features that are unique fingerprints of these three surgery types (Fig. 3c). For example, we observe tying in the second quartile of a procedure to be a key step of open appendectomy, consistent with completion of the division of the appendix. Other less obvious discriminative attributes emerge—such as variation of forceps use in the later segments—that exemplify how automated annotation can surface latent surgical attributes not obvious from superficial inspection. Overall, the multi-task model shows an ability to generalize across videos and surgical procedures to provide granular intraoperative details and a reference prototype for each surgery.

Our ultimate goal is to enable real-time feedback and tools for augmented decision-making. Thus, we optimized our model architecture to allow for real-time feedback and analysis requiring only a consumer-grade workstation. Unification of multiple vision tasks into a single multi-task architecture leads to a parameter efficient model which can run on a personal workstation with a single RTX 3090 GPU to generate spatial analytics every 0.08s and temporal action characterization every 0.33s. This rapid inference enables future integration into real-time surgical workflows with only minimal infrastructure requirements. 

\section{Quantifying Surgical Skill}
\label{sec:headings}

\begin{figure}[h]
    \centering
    \includegraphics[width=0.9\textwidth]{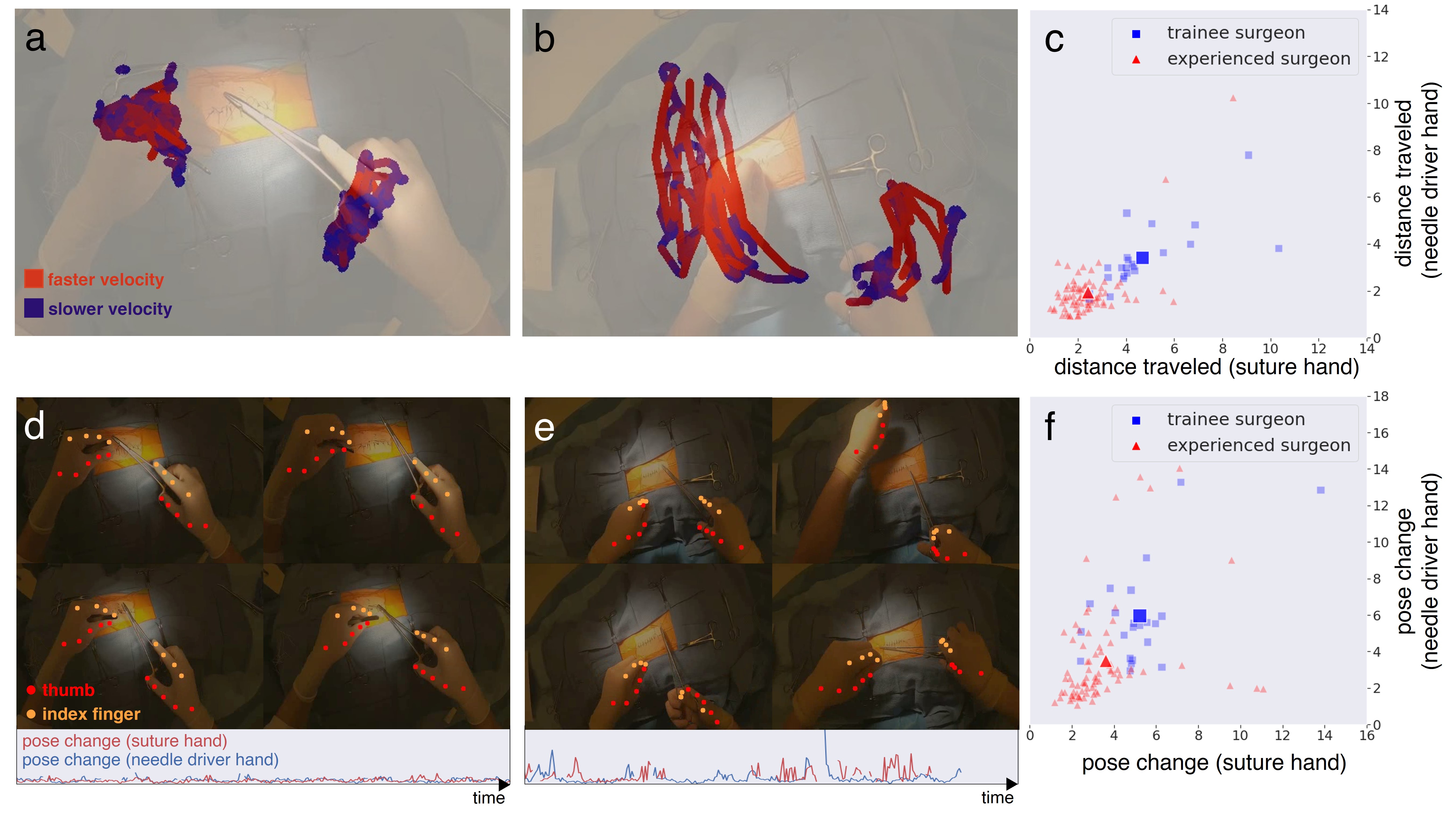}
    \caption{\textbf{Figure 4. Quantifying hand behavior in instrument-assisted surgical knot ties.} Hand paths for (a) an experienced surgeon and (b) a medical student. (c) Distances traveled (in units of hand-lengths) per instrument-assisted surgical tie for experienced versus trainee surgeons. Bright points represent group centroids. Series of frames with associated hand keypoints for (d) an experienced surgeon and (e) a medical student. Below each series are plots of pose change versus time. (f) Summed pose distance traveled per instrument-assisted tie for experienced versus trainee surgeons. See Extended Data Fig. 6 for calculation. Bright points represent group centroids. (c, f) are based on real-world surgeries, while (a-b, d-e) show simulated operating room images due to privacy considerations.}
\end{figure}

In addition to high-level procedure understanding, our multi-task computer vision model can evaluate surgeon maneuvers to study individual performance. AI can rapidly and quantitatively validate candidate metrics for skill by analyzing countless videos at sub-second resolution without the need for time-consuming and subjective manual annotations. Identifying attributes of individual skill allows surgeons to recognize opportunities for improving technique and patient outcomes \cite{stulberg, fesco}. 

As a concrete example, we investigated the motion of surgeon hands in instrument-assisted surgical knot tying (Fig. 4). We used our AI model to analyze videos recorded prospectively at Beth Israel Deaconess Medical Center via an IRB-approved protocol. Twenty-three open surgical procedures across varied general surgery specialties were collected, that included operative actions of 14 operators (medical students, postgraduate residents, surgical fellows, and attending surgeons). 
From these surgical videos, we compiled an evaluative dataset of 104 video clips of instrument-assisted surgical tying. We directly applied our trained multi-task neural network, without further training, to these real-world cases, thus demonstrating its ability to generalize to unseen environments and potentially shifted data distributions. Representative model outputs tracking hand motion are shown in Figs. 4a,b, where the trajectory of the red-blue line follows the path taken by the operator while performing suture tying. The color of each line segment represents instantaneous hand speed, with red colors indicating higher and blue indicating lower speeds. Overall, we observed that experienced surgeons’ hands (Fig. 4a) demonstrate more localized movement compared to trainees’ hands to accomplish the same task (Fig. 4b). Fig. 4c shows 104 acts of surgical knot tying across 14 operators, with each axis corresponding to the integrated distance traveled for each hand. Data representing experienced surgeons is clustered at shorter distances traveled, while data representing trainee surgeons is clustered at longer distances traveled, with larger spread, suggesting that within a surgical training program there is a range of skills for individual tasks. The data suggests that the distance each hand travels while knot tying is an indicator of surgical training, with shorter distances correlating with more experience. We observe that the prototypical experienced surgeon (bright red star) travels approximately two hand-lengths to complete a single tie, while the prototypical trainee (bright blue circle) travels approximately four hand-lengths. 

A more nuanced understanding of surgical technique is found by studying the changing relationship between the nine keypoints comprising thumb and index fingers, throughout time–a massively time-consuming measurement if done by manual annotation. Extended Data Fig. 6 shows how hand pose change between frames is easily quantified algorithmically. Representative sequences of model outputs for hand pose are shown in Figs. 4d,e. Qualitatively, we observe consistent positioning of the index finger and thumb in experienced surgeons, while trainees move between various hand poses. Line plots below Figs. 4d,e quantitatively show the change of hand pose as a function of time, normalized by hand size. While the experienced surgeons showed constant and low pose velocities, trainees showed numerous intermittent pose changes, which over time, leads to a significant amount of pose movement. The analysis easily scales to the dataset of 104 tying acts, and Fig. 4f plots integrated pose movement against level of surgical training for each hand. As with translational movement, we observe less pose movement in experienced surgeons compared to trainee surgeons. Given the dataset size, a robustness of our conclusions to outliers is shown in Extended Data. Fig 7.

The aggregate effects of hand translation and pose movement can be interpreted as elements of surgical skill such as decisiveness and economy of motion; more direct and decisive motions suggest adequate technical skill and intraoperative judgement. Current derivation of these metrics depends on qualitative human reviewers using surgical skill assessment tools such as Objective Structured Assessment of Technical Skills (OSATS), an assessment instrument which includes economy of motion and efficiency as indicators of technical skill \cite{martin}. Objective derivation of surgical skill assessment elements represents a step towards quantitative and automated evaluation of surgical skill that can facilitate skill improvement.

\section{Conclusion}
Surgeons train for years to achieve clinical intuition and technical mastery. Our AI model, which has been trained on a diverse international dataset of publicly-available videos, is a significant step towards parsing the elements of surgical procedures that lead to such mastery. We illustrated the use of our multi-task model outputs to generate example surgical signatures that define the prototypical flow of various surgery types, as well as to identify attributes of surgical skill through hand pose analysis on a prospective real-world dataset. These quantified attributes mirror those of current state-of-the-art surgical assessment tools requiring human qualitative evaluation. By understanding how the building blocks of actions, tools, and hand movements are characteristic of surgical procedures and levels of training, AI systems have the potential to provide post-hoc evaluation and real-time assistance. Such tools could synthesize the best techniques across numerous surgeons and countless surgical videos. Our work takes an important first step towards the development of artificial intelligence which can assist surgeons across diverse surgical procedures and operating scenes.


\printbibliography{} 

@article{nepogodiev,
  title={Global burden of postoperative death},
  author={Nepogodiev, Dmitri and Martin, Janet and Biccard, Bruce and Makupe, Alex and Bhangu, Aneel and Ademuyiwa, Adesoji and Adisa, Adewale Oluseye and Aguilera, Maria-Lorena and Chakrabortee, Sohini and Fitzgerald, J Edward and others},
  journal={The Lancet},
  volume={393},
  number={10170},
  pages={401},
  year={2019},
  publisher={Elsevier}
}

@article{sacks,
  title={Relationship between hospital performance on a patient satisfaction survey and surgical quality},
  author={Sacks, Greg D and Lawson, Elise H and Dawes, Aaron J and Russell, Marcia M and Maggard-Gibbons, Melinda and Zingmond, David S and Ko, Clifford Y},
  journal={JAMA surgery},
  volume={150},
  number={9},
  pages={858--864},
  year={2015},
  publisher={American Medical Association}
}

@article{birkmeyer,
  title={Surgical skill and complication rates after bariatric surgery},
  author={Birkmeyer, John D and Finks, Jonathan F and O'Reilly, Amanda and Oerline, Mary and Carlin, Arthur M and Nunn, Andre R and Dimick, Justin and Banerjee, Mousumi and Birkmeyer, Nancy JO},
  journal={New England Journal of Medicine},
  volume={369},
  number={15},
  pages={1434--1442},
  year={2013},
  publisher={Mass Medical Soc}
}

@article{martin,
  title={Objective structured assessment of technical skill (OSATS) for surgical residents},
  author={Martin, JA and Regehr, Glenn and Reznick, Richard and Macrae, Helen and Murnaghan, John and Hutchison, Carol and Brown, M},
  journal={Journal of British Surgery},
  volume={84},
  number={2},
  pages={273--278},
  year={1997},
  publisher={Oxford University Press}
}

@article{hashimoto,
  title={Artificial intelligence in surgery: promises and perils},
  author={Hashimoto, Daniel A and Rosman, Guy and Rus, Daniela and Meireles, Ozanan R},
  journal={Annals of surgery},
  volume={268},
  number={1},
  pages={70},
  year={2018},
  publisher={NIH Public Access}
}

@article{richards,
  title={A national review of the frequency of minimally invasive surgery among general surgery residents: assessment of ACGME case logs during 2 decades of general surgery resident training},
  author={Richards, Morgan K and McAteer, Jarod P and Drake, F Thurston and Goldin, Adam B and Khandelwal, Saurabh and Gow, Kenneth W},
  journal={JAMA surgery},
  volume={150},
  number={2},
  pages={169--172},
  year={2015},
  publisher={American Medical Association}
}

@article{esteva1,
  title={Deep learning-enabled medical computer vision},
  author={Esteva, Andre and Chou, Katherine and Yeung, Serena and Naik, Nikhil and Madani, Ali and Mottaghi, Ali and Liu, Yun and Topol, Eric and Dean, Jeff and Socher, Richard},
  journal={NPJ digital medicine},
  volume={4},
  number={1},
  pages={1--9},
  year={2021},
  publisher={Nature Publishing Group}
}

@article{esteva2,
  title={Dermatologist-level classification of skin cancer with deep neural networks},
  author={Esteva, Andre and Kuprel, Brett and Novoa, Roberto A and Ko, Justin and Swetter, Susan M and Blau, Helen M and Thrun, Sebastian},
  journal={nature},
  volume={542},
  number={7639},
  pages={115--118},
  year={2017},
  publisher={Nature Publishing Group}
}

@article{defauw,
  title={Clinically applicable deep learning for diagnosis and referral in retinal disease},
  author={De Fauw, Jeffrey and Ledsam, Joseph R and Romera-Paredes, Bernardino and Nikolov, Stanislav and Tomasev, Nenad and Blackwell, Sam and Askham, Harry and Glorot, Xavier and O’Donoghue, Brendan and Visentin, Daniel and others},
  journal={Nature medicine},
  volume={24},
  number={9},
  pages={1342--1350},
  year={2018},
  publisher={Nature Publishing Group}
}

@article{lundervold,
  title={An overview of deep learning in medical imaging focusing on MRI},
  author={Lundervold, Alexander Selvikv{\aa}g and Lundervold, Arvid},
  journal={Zeitschrift f{\"u}r Medizinische Physik},
  volume={29},
  number={2},
  pages={102--127},
  year={2019},
  publisher={Elsevier}
}

@article{ouyang,
  title={Video-based AI for beat-to-beat assessment of cardiac function},
  author={Ouyang, David and He, Bryan and Ghorbani, Amirata and Yuan, Neal and Ebinger, Joseph and Langlotz, Curtis P and Heidenreich, Paul A and Harrington, Robert A and Liang, David H and Ashley, Euan A and others},
  journal={Nature},
  volume={580},
  number={7802},
  pages={252--256},
  year={2020},
  publisher={Nature Publishing Group}
}

@inproceedings{zia,
  title={Automated assessment of surgical skills using frequency analysis},
  author={Zia, Aneeq and Sharma, Yachna and Bettadapura, Vinay and Sarin, Eric L and Clements, Mark A and Essa, Irfan},
  booktitle={International Conference on Medical Image Computing and Computer-Assisted Intervention},
  pages={430--438},
  year={2015},
  organization={Springer}
}

@inproceedings{carreira,
  title={Quo vadis, action recognition? a new model and the kinetics dataset},
  author={Carreira, Joao and Zisserman, Andrew},
  booktitle={proceedings of the IEEE Conference on Computer Vision and Pattern Recognition},
  pages={6299--6308},
  year={2017}
}

@article{chen,
  title={Estimating the quality of YouTube videos on pulmonary lobectomy},
  author={Chen, Zixuan and Zhu, Hongyu and Zhao, Weijun and Guo, Haixie and Zhou, Chengwei and Shen, Jianfei and Ye, Minhua},
  journal={Journal of thoracic disease},
  volume={11},
  number={9},
  pages={4000},
  year={2019},
  publisher={AME Publications}
}

@article{derakhshan,
  title={Assessing the educational quality of ‘YouTube’videos for facelifts},
  author={Derakhshan, Adeeb and Lee, Linda and Bhama, Prabhat and Barbarite, Eric and Shaye, David},
  journal={American journal of otolaryngology},
  volume={40},
  number={2},
  pages={156--159},
  year={2019},
  publisher={Elsevier}
}

@article{rapp,
  title={YouTube is the most frequently used educational video source for surgical preparation},
  author={Rapp, Allison K and Healy, Michael G and Charlton, Mary E and Keith, Jerrod N and Rosenbaum, Marcy E and Kapadia, Muneera R},
  journal={Journal of surgical education},
  volume={73},
  number={6},
  pages={1072--1076},
  year={2016},
  publisher={Elsevier}
}

@article{deSantibañes,
  title={Postoperative complications at a university hospital: is there a difference between patients operated by supervised residents vs. trained surgeons?},
  author={de Santiba{\~n}es, Martin and Alvarez, Fernando A and Sieling, Esteban and Vaccarezza, Hernan and de Santiba{\~n}es, Eduardo and Vaccaro, Carlos A},
  journal={Langenbeck's archives of surgery},
  volume={400},
  number={1},
  pages={77--82},
  year={2015},
  publisher={Springer}
}

@inproceedings{tran,
  title={A closer look at spatiotemporal convolutions for action recognition},
  author={Tran, Du and Wang, Heng and Torresani, Lorenzo and Ray, Jamie and LeCun, Yann and Paluri, Manohar},
  booktitle={Proceedings of the IEEE conference on Computer Vision and Pattern Recognition},
  pages={6450--6459},
  year={2018}
}

@inproceedings{lin,
  title={Focal loss for dense object detection},
  author={Lin, Tsung-Yi and Goyal, Priya and Girshick, Ross and He, Kaiming and Doll{\'a}r, Piotr},
  booktitle={Proceedings of the IEEE international conference on computer vision},
  pages={2980--2988},
  year={2017}
}

@inproceedings{sun,
  title={Deep high-resolution representation learning for human pose estimation},
  author={Sun, Ke and Xiao, Bin and Liu, Dong and Wang, Jingdong},
  booktitle={Proceedings of the IEEE/CVF Conference on Computer Vision and Pattern Recognition},
  pages={5693--5703},
  year={2019}
}

@inproceedings{bewley,
  title={Simple online and realtime tracking},
  author={Bewley, Alex and Ge, Zongyuan and Ott, Lionel and Ramos, Fabio and Upcroft, Ben},
  booktitle={2016 IEEE international conference on image processing (ICIP)},
  pages={3464--3468},
  year={2016},
  organization={IEEE}
}

@article{hashimoto2,
  title={Computer vision analysis of intraoperative video: automated recognition of operative steps in laparoscopic sleeve gastrectomy},
  author={Hashimoto, Daniel A and Rosman, Guy and Witkowski, Elan R and Stafford, Caitlin and Navarrete-Welton, Allison J and Rattner, David W and Lillemoe, Keith D and Rus, Daniela L and Meireles, Ozanan R},
  journal={Annals of surgery},
  volume={270},
  number={3},
  pages={414},
  year={2019},
  publisher={NIH Public Access}
}

@inproceedings{garcia,
  title={Toolnet: holistically-nested real-time segmentation of robotic surgical tools},
  author={Garcia-Peraza-Herrera, Luis C and Li, Wenqi and Fidon, Lucas and Gruijthuijsen, Caspar and Devreker, Alain and Attilakos, George and Deprest, Jan and Vander Poorten, Emmanuel and Stoyanov, Danail and Vercauteren, Tom and others},
  booktitle={2017 IEEE/RSJ International Conference on Intelligent Robots and Systems (IROS)},
  pages={5717--5722},
  year={2017},
  organization={IEEE}
}

@article{twinanda,
  title={Endonet: a deep architecture for recognition tasks on laparoscopic videos},
  author={Twinanda, Andru P and Shehata, Sherif and Mutter, Didier and Marescaux, Jacques and De Mathelin, Michel and Padoy, Nicolas},
  journal={IEEE transactions on medical imaging},
  volume={36},
  number={1},
  pages={86--97},
  year={2016},
  publisher={IEEE}
}

@article{ahmidi,
  title={A dataset and benchmarks for segmentation and recognition of gestures in robotic surgery},
  author={Ahmidi, Narges and Tao, Lingling and Sefati, Shahin and Gao, Yixin and Lea, Colin and Haro, Benjamin Bejar and Zappella, Luca and Khudanpur, Sanjeev and Vidal, Ren{\'e} and Hager, Gregory D},
  journal={IEEE Transactions on Biomedical Engineering},
  volume={64},
  number={9},
  pages={2025--2041},
  year={2017},
  publisher={IEEE}
}

@article{bodenreider,
  title={The unified medical language system (UMLS): integrating biomedical terminology},
  author={Bodenreider, Olivier},
  journal={Nucleic acids research},
  volume={32},
  number={suppl\_1},
  pages={D267--D270},
  year={2004},
  publisher={Oxford University Press}
}

@article{stulberg,
  title={Association between surgeon technical skills and patient outcomes},
  author={Stulberg, Jonah J and Huang, Reiping and Kreutzer, Lindsey and Ban, Kristen and Champagne, Bradley J and Steele, Scott R and Johnson, Julie K and Holl, Jane L and Greenberg, Caprice C and Bilimoria, Karl Y},
  journal={JAMA surgery},
  volume={155},
  number={10},
  pages={960--968},
  year={2020},
  publisher={American Medical Association}
}

@article{fesco,
  title={The effect of technical performance on patient outcomes in surgery},
  author={Fecso, Andras B and Szasz, Peter and Kerezov, Georgi and Grantcharov, Teodor P},
  journal={Annals of surgery},
  volume={265},
  number={3},
  pages={492--501},
  year={2017},
  publisher={Wolters Kluwer}
}

@article{richardson,
  title={Beautiful soup documentation},
  author={Richardson, Leonard},
  journal={Dosegljivo: https://www. crummy. com/software/BeautifulSoup/bs4/doc/.[Dostopano: 7. 7. 2018]},
  year={2007}
}

@article{society,
    author = "Society, S. C. H",
    title = "List of surgical procedure",
    pages="1–-44",
    year="2014",
}

@inproceedings{he,
  title={Deep residual learning for image recognition},
  author={He, Kaiming and Zhang, Xiangyu and Ren, Shaoqing and Sun, Jian},
  booktitle={Proceedings of the IEEE conference on computer vision and pattern recognition},
  pages={770--778},
  year={2016}
}

@article{russakovsky,
  title={Imagenet large scale visual recognition challenge},
  author={Russakovsky, Olga and Deng, Jia and Su, Hao and Krause, Jonathan and Satheesh, Sanjeev and Ma, Sean and Huang, Zhiheng and Karpathy, Andrej and Khosla, Aditya and Bernstein, Michael and others},
  journal={International journal of computer vision},
  volume={115},
  number={3},
  pages={211--252},
  year={2015},
  publisher={Springer}
}

@article{kingma,
  title={Adam: A method for stochastic optimization},
  author={Kingma, Diederik P and Ba, Jimmy},
  journal={arXiv preprint arXiv:1412.6980},
  year={2014}
}

@article{ren,
  title={Faster r-cnn: Towards real-time object detection with region proposal networks},
  author={Ren, Shaoqing and He, Kaiming and Girshick, Ross and Sun, Jian},
  journal={Advances in neural information processing systems},
  volume={28},
  pages={91--99},
  year={2015}
}

@inproceedings{simon,
  title={Hand keypoint detection in single images using multiview bootstrapping},
  author={Simon, Tomas and Joo, Hanbyul and Matthews, Iain and Sheikh, Yaser},
  booktitle={Proceedings of the IEEE conference on Computer Vision and Pattern Recognition},
  pages={1145--1153},
  year={2017}
}

@inproceedings{selvaraju,
  title={Grad-cam: Visual explanations from deep networks via gradient-based localization},
  author={Selvaraju, Ramprasaath R and Cogswell, Michael and Das, Abhishek and Vedantam, Ramakrishna and Parikh, Devi and Batra, Dhruv},
  booktitle={Proceedings of the IEEE international conference on computer vision},
  pages={618--626},
  year={2017}
}

\section{Methods}
\subsection{AVOS dataset}
All YouTube videos in AVOS were collected via web-scraping using BeautifulSoup, Selenium with Chromedriver, and youtube-dl \cite{richardson}. Videos were collected from 23 surgical procedure types, which were curated from a list of insured surgical procedures identified by the Southern Cross Health Society \cite{society}. These procedure types were selected by surgeons to span a range of common procedures in breast, gastrointestinal, and head-and-neck surgeries. Briefly, search results generated from the 23 surgical procedure queries were retrieved from YouTube via BeautifulSoup and Selenium. Each search term produced 400-600 results, which when consolidated, resulted in 9197 candidate surgical videos. Of these candidate surgical videos, many contained content outside the scope of our study, including laparoscopic surgeries, slide decks, personal narratives, and other non-open-surgical content. Each of these 9197 YouTube videos was therefore manually inspected for open surgical content, and 1997 videos were identified, comprising the Annotated Videos of Open Surgery (AVOS) dataset. Each of the 1997 surgical videos was also collected with available metadata annotations including views at time of collection, video duration, ‘like’ percentage, original search terms used to identify the video, and UMLS tags \cite{bodenreider}. We note that a specific YouTube ID may be associated with numerous search terms, and therefore, in the metadata, each video is associated with all search terms under which it appeared, as well as the numerical identifiers indicating where in the list of search results it appeared. When available (approx. 48\% of videos), the country of origin of the uploader’s account was included. In plotting the distribution of video lengths via box-plot (Fig. 1b), the center line represents the data median, box limits represent upper and lower quartiles, whiskers extend to represent 1.5x the interquartile range, and points represent outliers. We note that the AVOS dataset is inherently biased, as it presumably often contains superlative content intended for didactic or marketing purposes. The dataset will be made available upon reasonable request from the authors, and the training code, inference code, and trained models are available at (https://github.com/yeung-lab/spatiotemporal-open-surgery).

From this larger dataset, 343 open surgery videos were temporally annotated throughout at a per-second resolution with three actions (cutting, suturing, and tying) and a background class. 3430 frames were extracted, ten uniformly sampled per video, and densely labeled with spatial annotations including hand and tool (electocautery, needle driver, forceps) bounding boxes. A subset of these was annotated with 21 hand keypoints. Dense annotations were obtained using a custom annotator. Trained research assistants performed the labeling under the supervision of a board-certified surgeon, with each annotation verified by a second research assistant. 288 videos were used for training/validation and 55 videos were used for testing of the model. Additional statistics of the spatial and temporal annotations can be found in Extended Data Table 1.

\subsection{Training a multi-task model for detailed scene understanding}
Although many deep learning models exist for object detection, keypoint recognition, and action identification, few exist that can perform these objectives simultaneously. It is challenging to develop a training strategy to simultaneously use video data for action recognition and image data for object and keypoint identification. We adopted an alternating task training strategy, alternating forward passes of a short video clip (5 seconds, 64 frames) for training the action recognition head, with forward passes of two image frames for training the object detection heads. Image data was randomly sampled (i.e. not necessarily from the same video) while video data consisted of sequential frames from a 5 second video clip. When training batches consisted of video data, a standard cross entropy loss was backpropagated from the action head, and when training batches consisted of image data, a focal loss (hyperparameter gamma = 2) was backpropagated from each of the detection heads \cite{lin}. These two losses were summed together, with a weighting coefficient on the action loss of 0.0033, to create the overall model training loss. 

This weighting coefficient was chosen such that the training process is not dominated by maximizing accuracy for the action head. Data augmentation strategies were employed, including image flipping, scaling, rotation, sheering, blur, sharpening, and cropping. The shared ResNet \cite{he} encoder and object detection heads were pretrained using ImageNet \cite{russakovsky}, while the action head was randomly initialized. Models were trained for approximately 120 epochs, using the Adam \cite{kingma} optimizer with an initial learning rate of 1E-5 and a scheduler to reduce the rate by a factor of 2 every time the loss plateaus for 4 epochs. The model which performed best on the validation set was selected. Model training was performed using Google Cloud Platform. One NVIDIA Tesla A100 was used to train each model, and inference was performed either on a Tesla A100 or an NVIDIA GeForce RTX 3090 GPU. Training and inference were done using PyTorch. For action segmentation accuracy, precision and recall metrics were used for fair measurement of performance on the unbalanced distribution of surgical actions (Extended Data Table 2). For hand and tool detection, performance was measured via mean average precision at IoU = 0.5, which is common for object detection. 

Due to its modular nature, the hand pose keypoint branch was trained separately using one NVIDIA Tesla V100 GPU on Amazon Web Services. To increase our training dataset of hand keypoints, additional pseudolabel keypoints were bootstrapped from AVOS. To do this, a Deep High Resolution Net-32 (DHRN-32) model \cite{sun} was trained on the existing AVOS keypoints, and a Faster-RCNN model \cite{ren} was trained on AVOS hand bounding boxes. Using these initial models, we performed inference on the five frames before and after each human-labeled frame in our dataset, thereby creating a 10x increase in the number of annotations via pseudolabels on which to train our final keypoints model. The final keypoints model utilizes the DHRN-48 architecture, which we first pretrained on hand keypoint data from the CMU Panoptic Hands dataset \cite{sun, simon}, and then trained on ground truth AVOS keypoints as well as bootstrapped AVOS pseudolabels. The Probability of Correct Keypoint (PCK) was calculated by comparing the proximity of each predicted keypoint to the ground truth keypoint, versus the characteristic length of the detected hand bounding box; the probability of correct keypoint is therefore the fraction of predicted keypoints “close enough” to the ground truth keypoint, normalized by the size of the detected hand.

Inference is performed slightly differently than training; batches of four frames at a time are sequentially extracted from each video and passed into the neural network. After the first 16 batches during inference have been analyzed, the action head begins operating on the 64 stored activations from those batches, and continues operating on batches of 64 frames while stepping forward in the video, 4 frames at a time.

Techniques such as Grad-CAM \cite{selvaraju} are often used to generate saliency maps to interpret model predictions, but these are typically designed for the task of image classification and not for the more complex tasks of object detection and temporal segmentation. Nevertheless, we performed occluded detection experiments (Extended Data Fig. 3) by occluding different regions of the image and subsequently performing hand and tool detection on the partially occluded image. Colorful pixels represent regions when, when occluded by a black box, significantly depress detection confidence. We show that the regions important to confident object detection correspond to regions near key objects, suggesting that the model is detecting hands and tools based on reasonable object features.

\subsection{Surgical signatures}
By using the general descriptive building blocks of action, tool, and hand detections, we aimed to build a model capable of detailed scene understanding that can generalize across procedure types and visual scenes. Starting with AVOS, we curated an analysis dataset composed of appendectomies, pilonidal cystectomies, and thyroidectomies on which the model had not previously been exposed. Videos were filtered from AVOS to ensure that mostly complete procedures were selected for analysis. Videos for each surgery type were filtered based on a specific substrings appearing (1) in the UMLS keywords, and (2) in the original youtube search query keywords. To filter out overly edited procedures, videos were limited to those spanning 2-30 minutes. In certain cases, manual review was needed to ensure a level of homogeneity of procedures, (i.e. not just opening incisions, or advertisements for surgical tools). 

\begin{center}
\begin{tabular}{| p{3cm} | p{2.3cm} | p{3cm} | p{3cm} | p{3cm} | p{3cm} |} 
\hline
Surgical specialty & UMLS keyword & Search term keyword & Surgery time range & Additional procedure \\
\hline
Appendectomy & "append" & "append" & [2min,30min] & In title: "append" \\ 
\hline
Pilonidal Cystectomy & "flap" & "pilonidal" & [2min,30min]" & In title: (“karydakis”) OR \newline ((“pilon” OR “perin”) AND “flap”) \\
\hline
Thyroidectomy & "thyroid" & "thyroid" & [2min,30min] & In title: Manual confirmation of opening and closing scenes \\ 
\hline
\end{tabular}
\end{center}

In Fig. 3a,b, inference was performed on the collected specialty videos and the observed sequence of actions and tool presence were plotted. Background action segments were excised to increase homogeneity between procedures, and inference was performed at a five-second temporal resolution. To create surgical prototypes, procedure actions and tool counts were averaged for each quartile of the procedure, and a moving average filter was applied. For linear projection of surgeries into 2D space each surgery was parameterized into 30 interpretable features (according to Extended Data Fig. 4), and linear discriminant analysis (LDA) was performed via scikit-learn. Prior to LDA, each of the 30 features was normalized to zero mean and unit standard deviation. Therefore, the features with highest weights correspond to the most important features used for distinguishing different surgical characteristics.

\subsection{Collection of surgical videos in the operating room and simulated operating room}
Videos were collected using GoPro Hero 8 and 9 compact action cameras mounted to anesthesia poles and/or on the surgeon’s head. All videos were recorded at 1080p resolution and 30fps with standard GoPro settings. Positions of cameras were standardized to improve field of view and consistency in angle across procedures. Audio was not recorded. Basic post-processing of the collected video included cropping to the surgical site and reducing overexposure. In the small fraction of video segments with potentially identifiable features, these features were manually blurred. All videos were collected at Beth Israel Deaconess Medical Center (BIDMC), and this study was approved by the Institutional Review Board of BIDMC. For the qualitative analyses in Fig. 4a,b,d,e, due to privacy considerations, simulation videos were utilized. Consent was obtained from all participants in the study.

\subsection{Quantifying surgical skill}
The studied surgical operators were first divided into two categories: trainee surgeons (medical students, postgraduate residents) and experienced surgeons (surgical fellows, attending surgeons).
To understand factors that contribute to surgical skill, we analyzed the hands of individual surgeons to calculate kinematic parameters (e.g. distance traveled, velocity, acceleration). These metrics were calculated using the hand bounding box centroid as reference. The hand keypoint model allowed for detailed description of individual joint articulation within the hand, and we used these keypoints to measure the steadiness and variation of pose of an operating surgeon. In these analyses, we used nine hand keypoints, four for each of the thumb and index fingers, and one for the palm, to describe surgical skill.

Our workflow for analysis of surgical skill begins with video inference using the multi-task model. Tracking of each hand in the video is then performed with Simple Online and Realtime Tracking algorithm \cite{bewley}. At this point, short snippets (generally 10-20s long) of surgeons performing instrument-assisted surgical ties were identified by trained medical professionals. These segments were defined to begin at the first frame at which the suture crosses the needle driver, and to end at the frame where the suture is tied for the last time, after a series of knots (i.e. before the suture is cut). As different surgeons throw different numbers of knots, we also counted the number of knots tied (which generally ranges from 3-7), and divided by this to normalize certain metrics. To control for differences in camera positioning and zoom, velocity and distance traveled were normalized by the average hand bounding box size throughout the video clip. For calculation of hand velocity, acceleration, and jerk, the centroid of each bounding box is used to describe the location of the hand. Velocities may be first interpreted in units of (pixels/frame). To account for different levels of zoom, this unit is then normalized by the average size ((height+width)/2) of the operator hand integrated throughout the clip, and then normalized by the frame rate (frames/s) to produce a final velocity unit of $s^{-1}$. A similar process is used to calculate the amount of pose variation through keypoints, where distances are originally calculated in pixels, but normalized by dividing by the size of the operator’s hand. The exact calculation of hand pose change is detailed in Extended Data Fig. 5.

\section{Data availability}
Data will be made available by request upon publication.

\section{Code availability}
The code for training and running inference with the multi-task model will be made available upon publication.

\section*{Acknowledgments}
E.D.G. was funded partially by the National Library of Medicine under grant T15LM007033. Computational credits were generously contributed by Google Cloud Platform and Amazon Web Services through Stanford’s Institute for Human-Centered Artificial Intelligence. We thank Daniel Copeland and Michael Zhang for their contributions to dataset curation.

\textbf{Author contributions} All authors helped in dataset curation and labeling. E.D.G., K.K.P., W.L., S.R., R.M., and S.Y. designed the AI algorithm. Y.Z, C.K., J.C., H.W., M.D. and G.B. helped identify clinically meaningful descriptors of open surgery and surgical skill. E.D.G., K.K.P., and Y.Z. wrote the initial paper draft, and all authors contributed to paper revision and editing. S.Y. and G.B. conceptualized and supervised the research. 

\textbf{Competing interests} The authors declare no competing interests.

\newpage
\begin{figure}[h]
    \centering
    \includegraphics[width=0.6\textwidth]{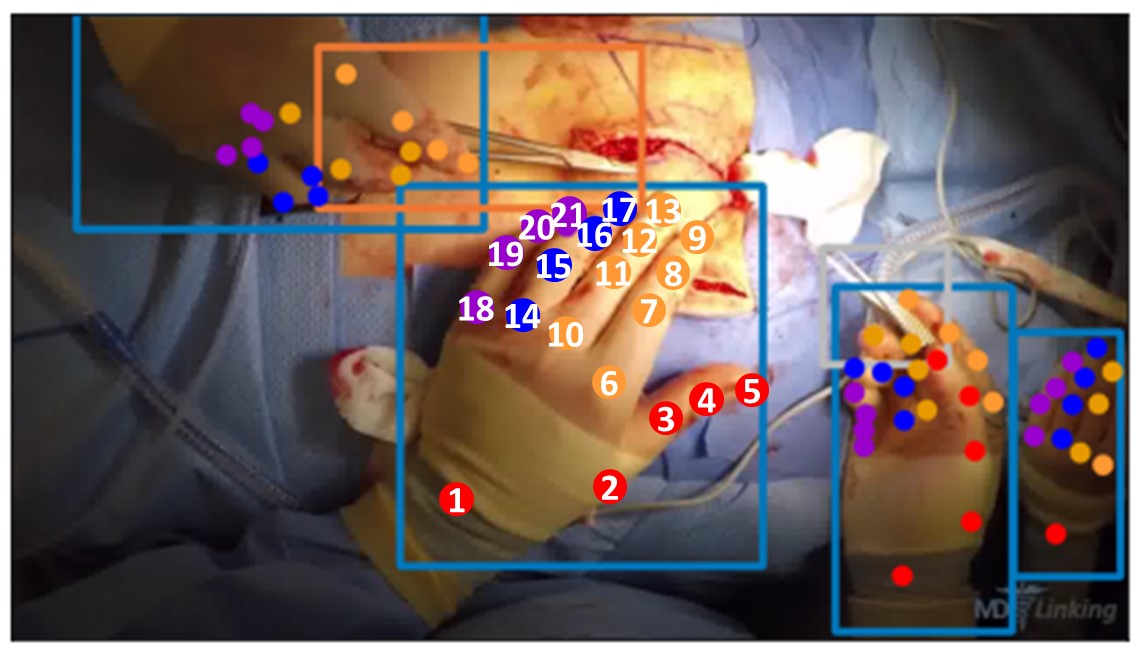}
    \caption{\textbf{Extended Data Figure 1. Counting 21 hand keypoints.} Example annotation identifying the 21 hand keypoints.}
\end{figure}

\newpage
\begin{figure}[h]
    \centering
    \includegraphics[width=0.8\textwidth]{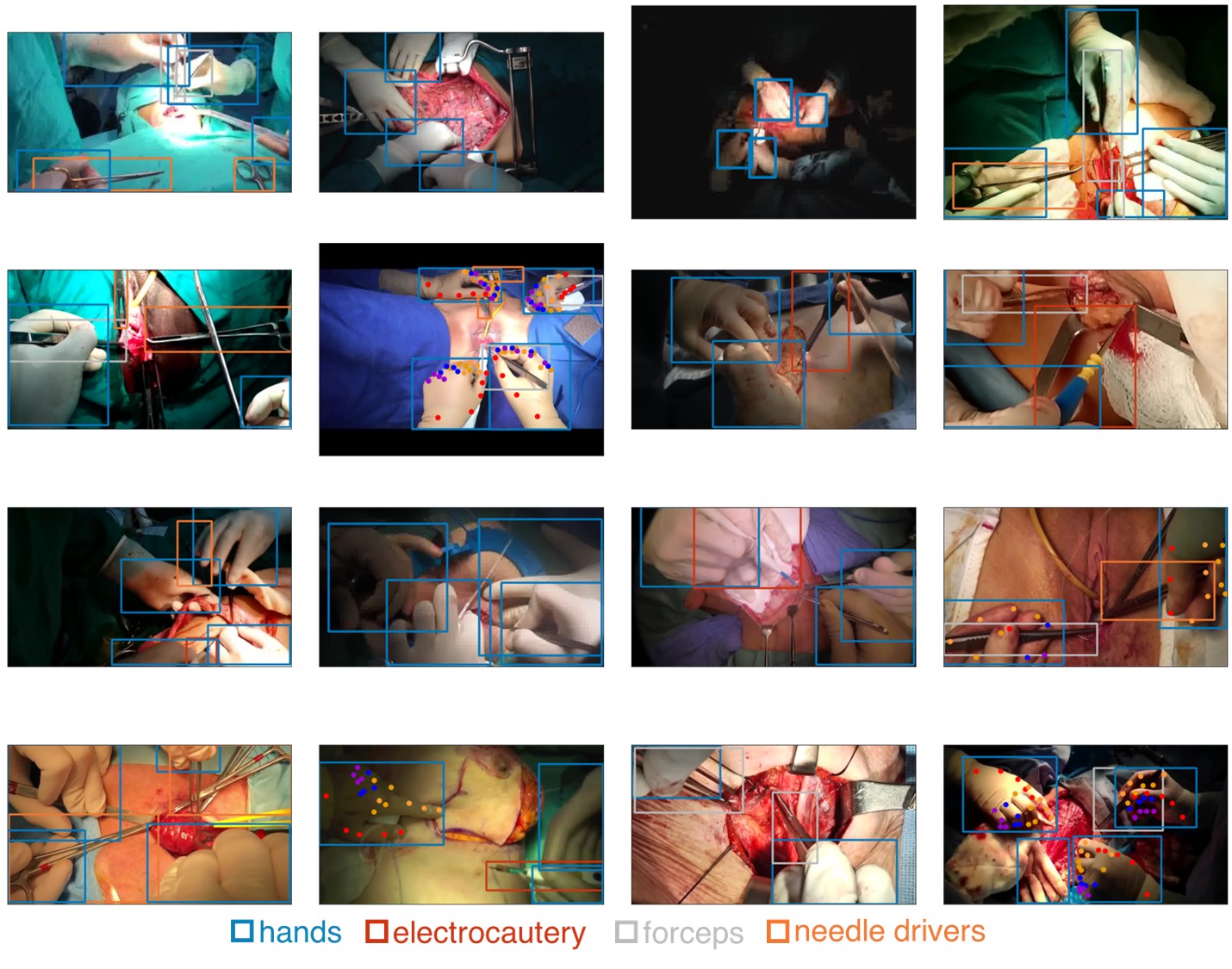}
    \caption{\textbf{Extended Data Figure 2. Diversity of AVOS dataset.} Selected images with corresponding bounding box annotations from the training dataset. A subset of the images additionally has hand keypoint annotations.}
\end{figure}

\newpage
\begin{figure}[h]
    \centering
    \includegraphics[width=0.8\textwidth]{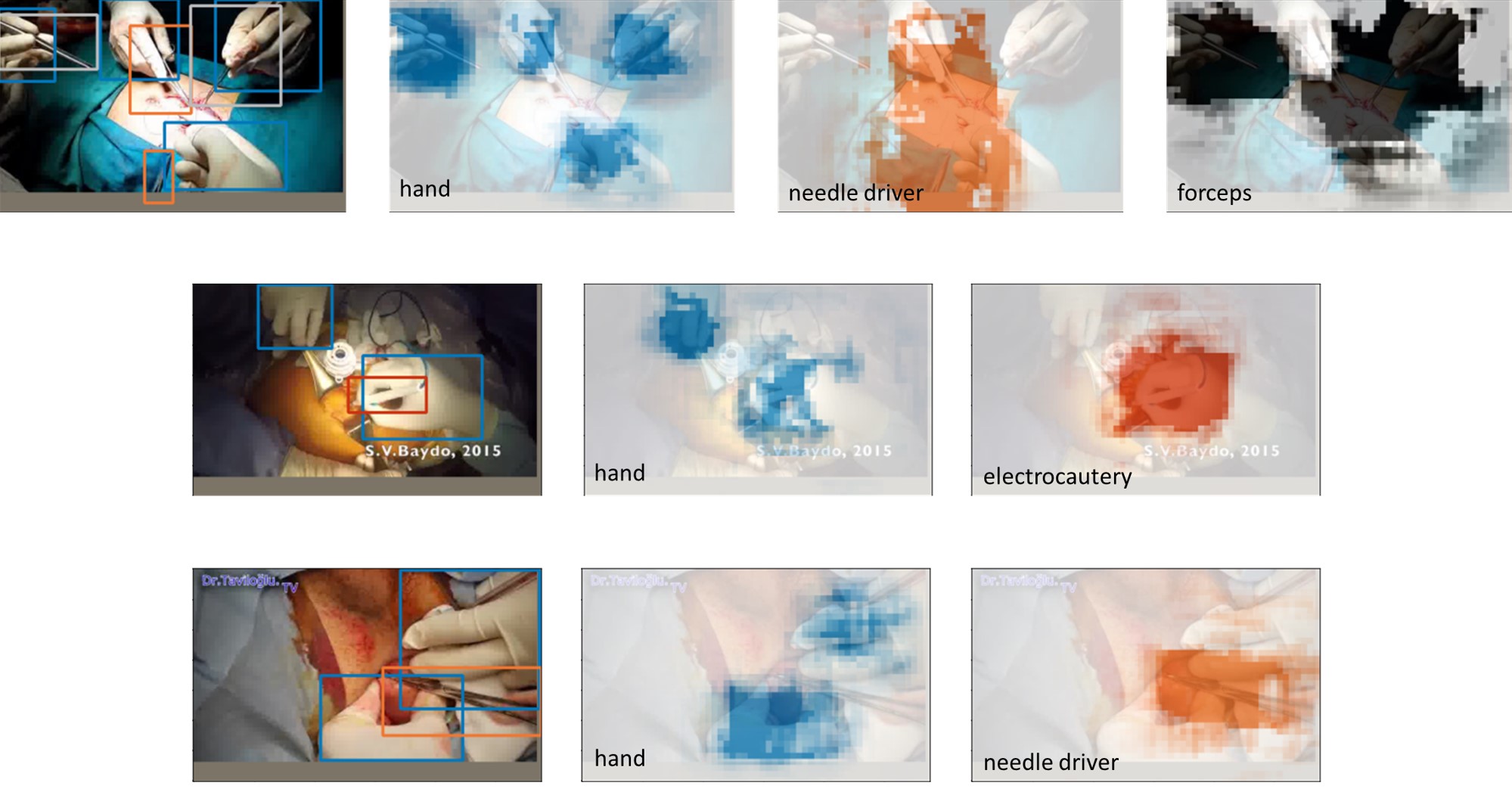}
    \caption{\textbf{Extended Data Figure 3. Interpretability of object detection heads.} For three different images, spatial analysis of important areas for detection of hands and tools. Colored areas indicate that proximal regions were key for confident detections of the corresponding hand or instrument.}
\end{figure}

\newpage
\begin{figure}[h]
    \centering
    \includegraphics[width=0.8\textwidth]{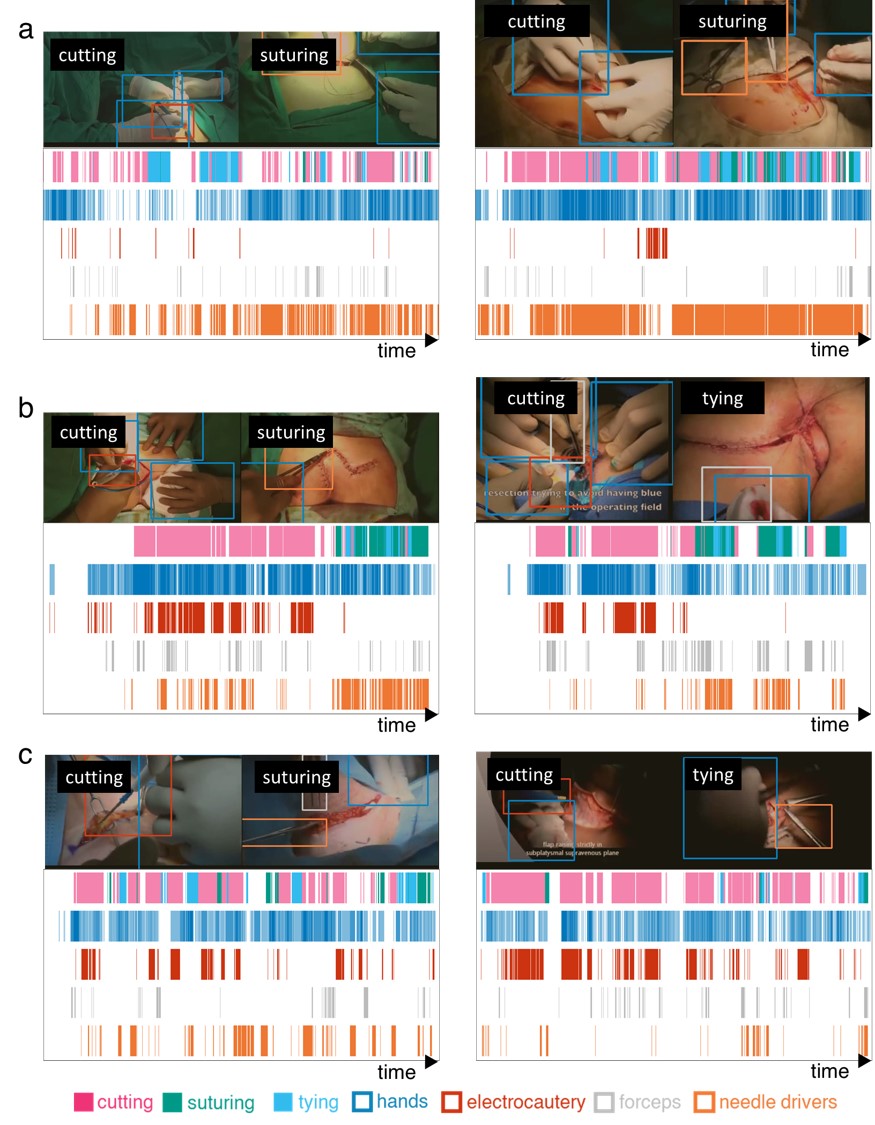}
    \caption{\textbf{Extended Data Figure 4. Examples of model inference for action, hand, and tool detections on open surgery procedures.} Representative model outputs for (a) appendectomy, (b) pilonidal cystectomy, and (c) thyroidectomy procedures. For each of the six example procedures, left picture is from opening incision, while right picture is from closure.}
\end{figure}

\newpage
\begin{figure}[h]
    \centering
    \includegraphics[width=0.8\textwidth]{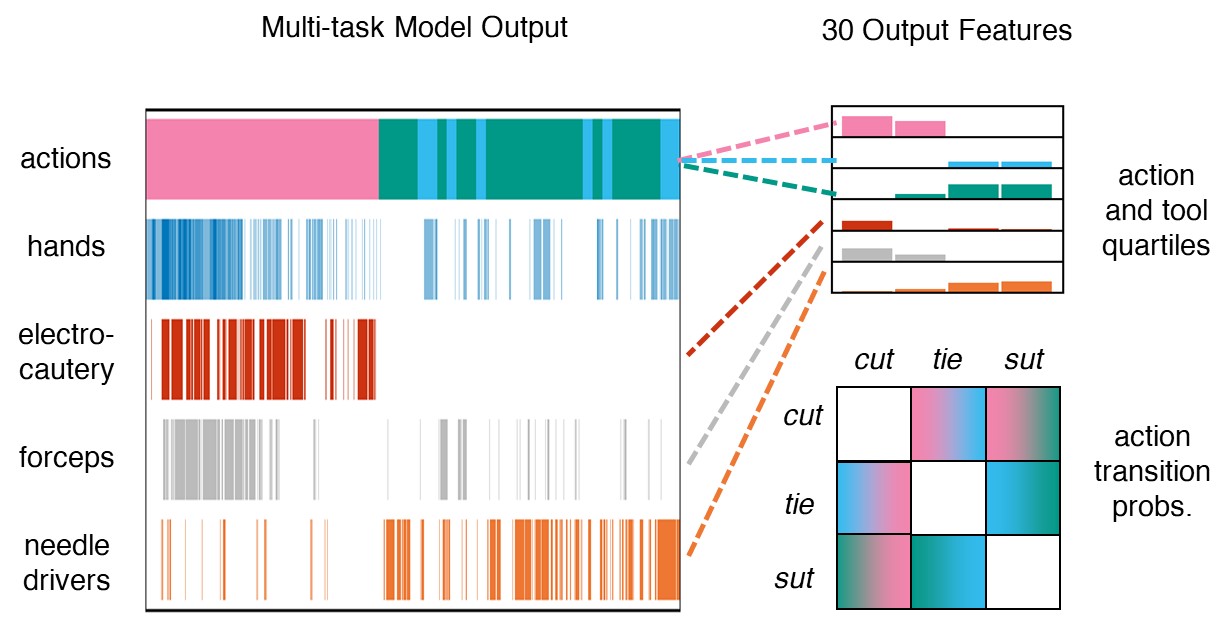}
    \caption{\textbf{Extended Data Figure 5. Parameterization of a single surgical video into 30 interpretable features for linear discriminant analysis.} 24 features (top right) related to the normalized amount of action and tool use through the procedure, separated into quartiles of time, and 6 features (bottom right) related to the probability of transitioning between surgical actions.}
\end{figure}

\newpage
\begin{figure}[h]
    \centering
    \includegraphics[width=0.8\textwidth]{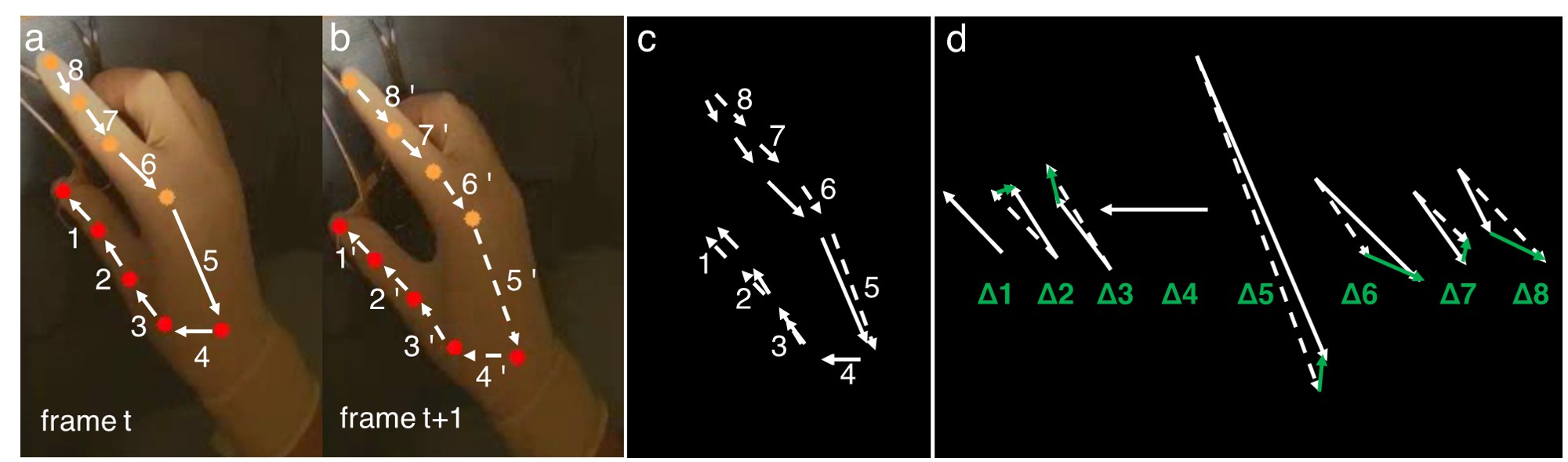}
    \caption{\textbf{Extended Data Figure 6. Calculating differences between hand poses.} Each hand pose (a, b) is parameterized into eight vectors representing distances between sequential keypoints. (c) Next, we compare the relative descriptive vectors between pose at frame t (solid arrows) and frame t+1 (dotted arrows). (d) To calculate total hand pose change between frames, the summed L1 distance between corresponding vectors is used. Analyses using this calculation are shown in main text Figs 4d-f.}
\end{figure}

\newpage
\begin{figure}[h]
    \centering
    \includegraphics[width=0.95\textwidth]{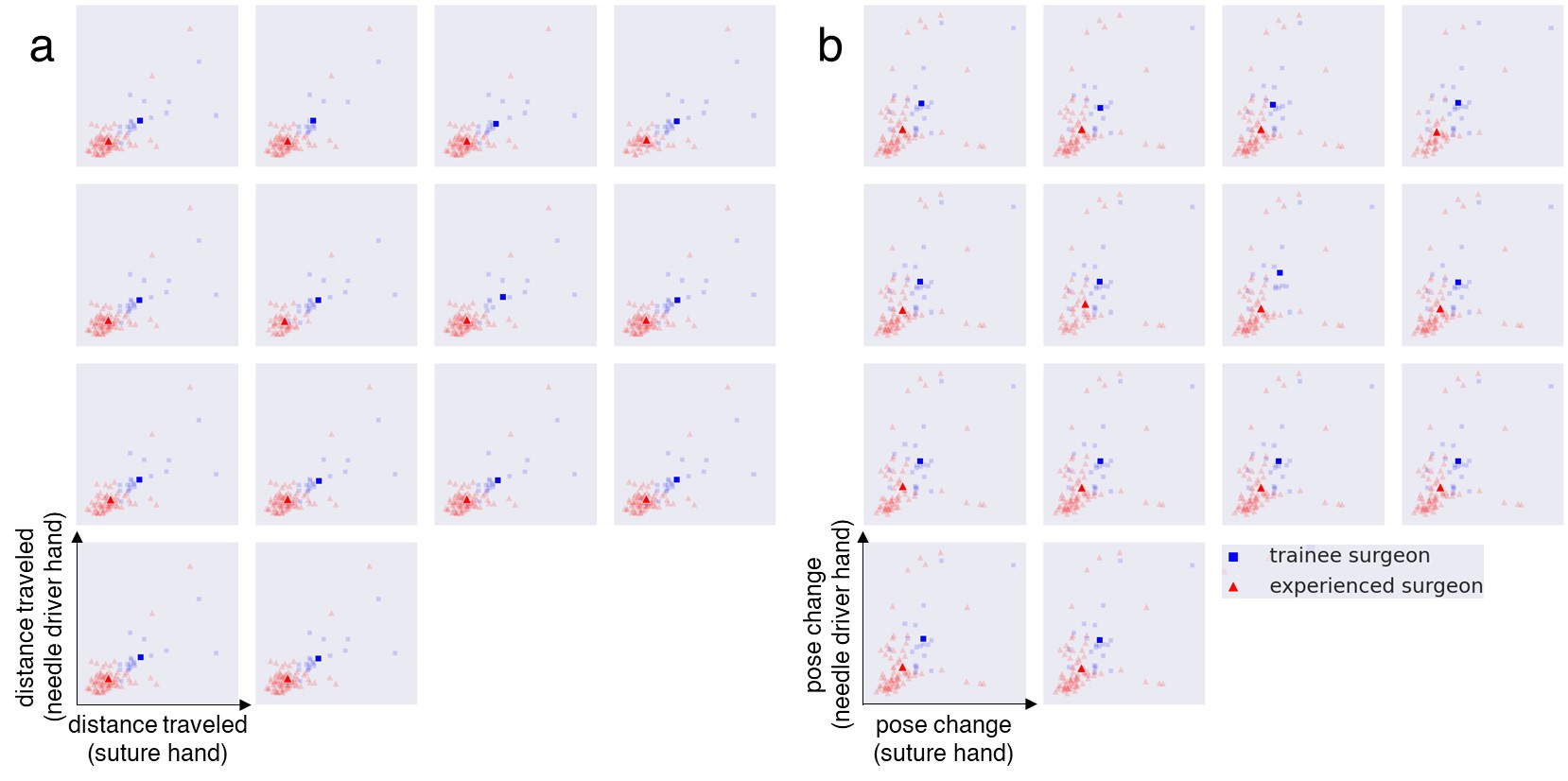}
    \caption{\textbf{Extended Data Figure 7. Leave-one-out cross-validation of skill centroids.} Each plot represents analyses shown in Figs. 4c,f, minus an operator in the dataset. (a) Distribution of hand distances traveled versus skill level, minus one operator per plot. (b) Distribution of hand pose change versus skill level, minus one operator per plot. We observe no significant shift in the centroid of each class with the removal of individual operators.}
\end{figure}

\newpage
\begin{figure}[h]
    \centering
    \includegraphics[width=0.95\textwidth]{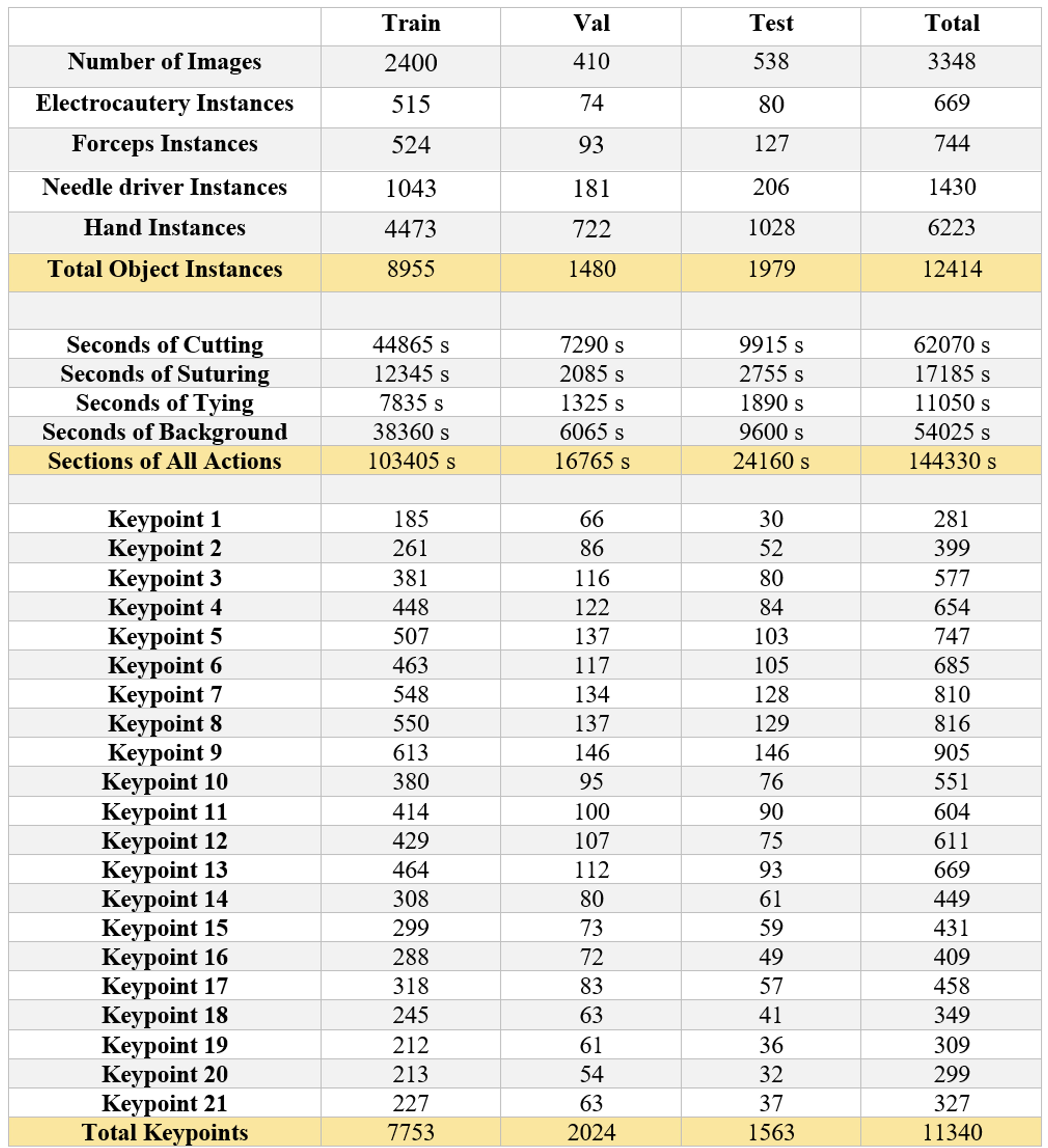}
    \caption{\textbf{Extended Data Table 1. Detailed scene annotation statistics of AVOS.} (Top) Number of object instances. (Middle) Seconds of annotated surgical actions and background class. (Bottom) Number of hand keypoints.}
\end{figure}

\newpage
\begin{figure}[h]
    \centering
    \includegraphics[width=0.95\textwidth]{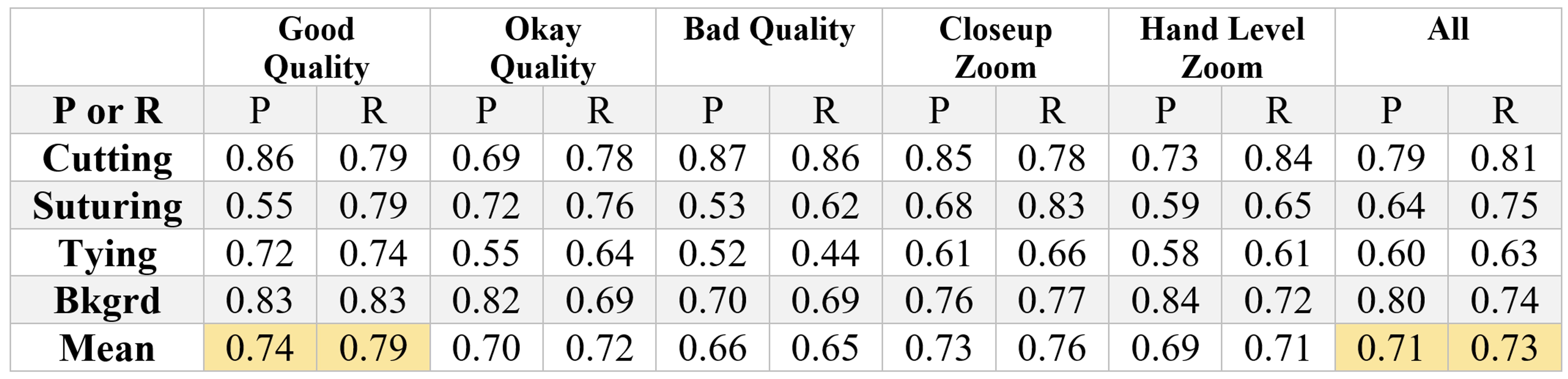}
    \caption{\textbf{Extended Data Table 2. Detailed breakdown of precision and recall on test data.} Precision/recall of actions on videos of different quality and different zooms levels. Highlighted numbers described in main text.}
\end{figure}

\begin{figure}[h]
    \centering
    \includegraphics[width=0.95\textwidth]{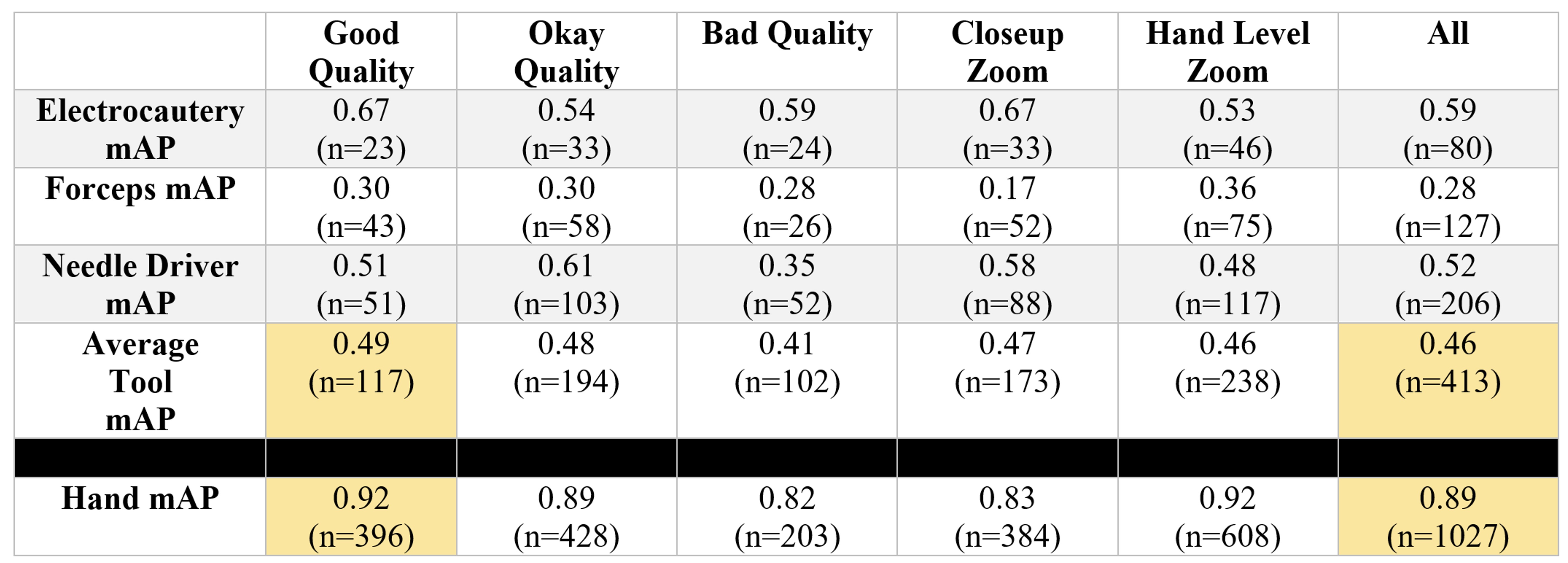}
    \caption{\textbf{Extended Data Table 3. Detailed breakdown of mAP on test data.} mAP of tools and hands on images of different quality and zooms levels using an IOU threshold of 0.5. Highlighted numbers described in the main text.}
\end{figure}

\newpage
\begin{figure}[h]
    \centering
    \includegraphics[width=0.95\textwidth]{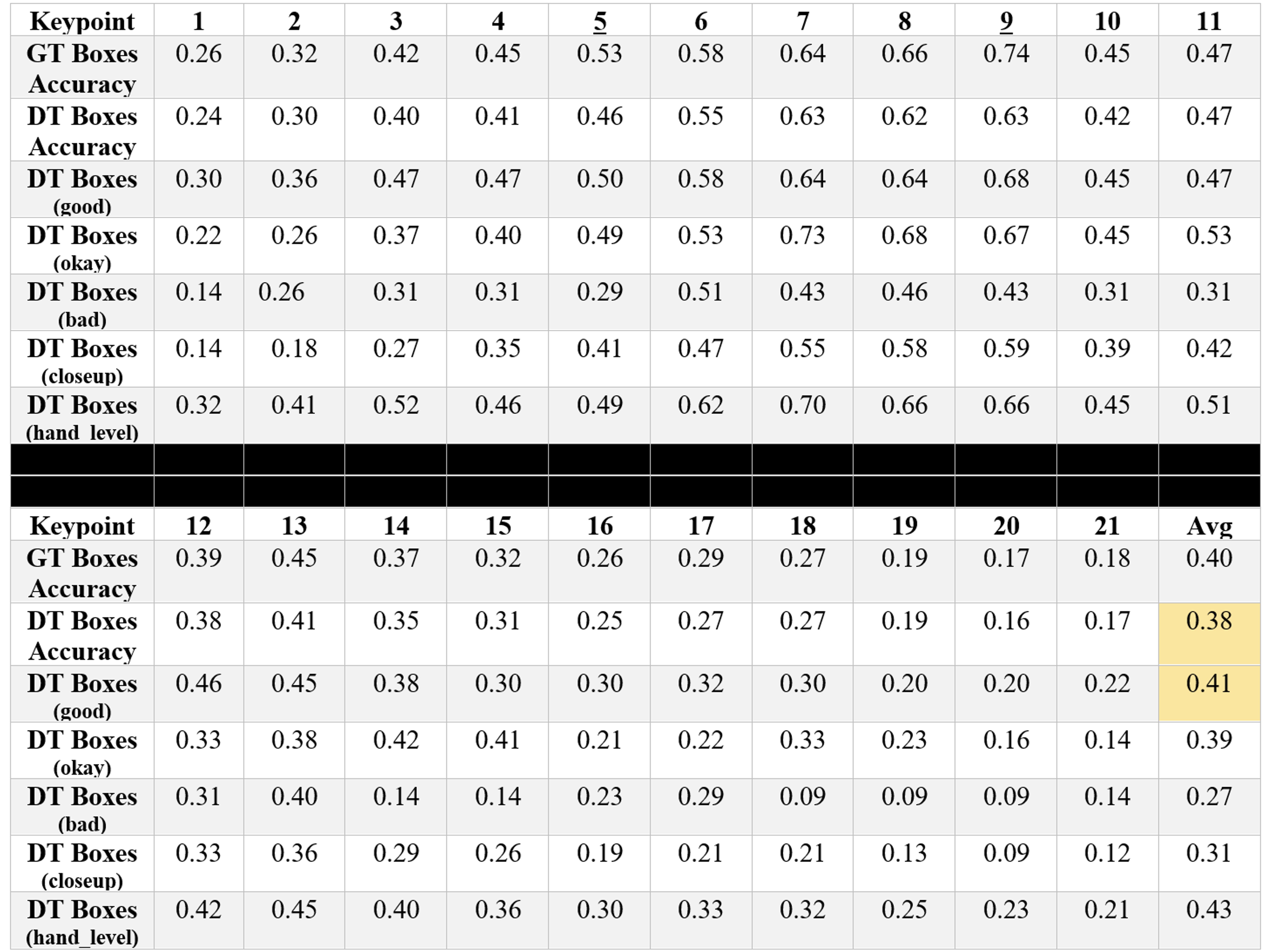}
    \caption{\textbf{Extended Data Table 4. Detailed breakdown of probability of correct keypoint (PCK) on test data.} PCK of 21 hand keypoints on test images. Highlighted numbers discussed in the main text. Each row represents a different type of analysis; GT = keypoint detection performed using ground truth hand boxes, DT = keypoint detection performed using detected hand boxes output by the multi-task neural network, and parentheticals represent subsets of data with specific data qualities.}
\end{figure}

\end{document}